\documentclass[conference]{IEEEtran}

\usepackage{makecell}
\usepackage{amsmath,amssymb,amsfonts}
\usepackage{xcolor}
\usepackage{ragged2e}
\usepackage{algorithm}
\usepackage{algcompatible}

\usepackage{graphicx}
\usepackage{textcomp}
\usepackage{rotating}
\usepackage{balance}
\usepackage{multirow}
\usepackage{gensymb}
\usepackage{array,booktabs,longtable,tabularx,tabulary, subfig}
\usepackage{lipsum, balance}

\usepackage[pscoord]{eso-pic}
\usepackage{hyphenat}

\usepackage{threeparttable, booktabs, url, eqparbox, xspace, setspace, tabularx}
\usepackage[colorlinks,urlcolor=blue,linkcolor=blue,citecolor=blue]{hyperref}
\makeatletter
\DeclareRobustCommand\onedot{\futurelet\@let@token\@onedot}
\def\@onedot{\ifx\@let@token.\else.\null\fi\xspace}

\def\etal{\emph{et al}\onedot}

\hypersetup{
    colorlinks = true,
    linkcolor  = black,
    filecolor = black,      
    urlcolor = black,
    citecolor=black,
} 
\makeatletter
 \let\old@ps@headings\ps@headings
 \let\old@ps@IEEEtitlepagestyle\ps@IEEEtitlepagestyle
 \def\confheader#1{%
 \def\ps@IEEEtitlepagestyle{%
 \old@ps@IEEEtitlepagestyle%
 \def\@oddhead{\strut\hfill#1\hfill\strut}%
 \def\@evenhead{\strut\hfill#1\hfill\strut}%
 }%
 \ps@headings%
 }
 \makeatother
\begin{document}

\title{Effects of Real-Life Traffic Sign Alteration on YOLOv7- an Object Recognition Model}

\author{
    \IEEEauthorblockN{%
        Farhin Farhad Riya,
        Shahinul Hoque,
        Md Saif Hassan Onim,
        Edward Michaud,
        Edmon Begoli, and
        Jinyuan Stella Sun}

    \IEEEauthorblockA{%
        Department of Electrical Engineering and Computer Science,\\
        University of Tennessee, Knoxville}
    }
    
\maketitle

\begin{abstract}
The widespread adoption of Image Processing has propelled Object Recognition (OR) models into essential roles across various applications, demonstrating the power of AI and enabling crucial services. Among the applications, traffic sign recognition stands out as a popular research topic, given its critical significance in the development of autonomous vehicles. Despite their significance, real-world challenges, such as alterations to traffic signs, can negatively impact the performance of OR models. This study investigates the influence of altered traffic signs on the accuracy and effectiveness of object recognition, employing a publicly available dataset to introduce alterations in shape, color, content, visibility, angles and background. Focusing on the YOLOv7 (You Only Look Once) model, the study demonstrates a notable decline in detection and classification accuracy when confronted with traffic signs in unusual conditions including the altered traffic signs. Notably, the alterations explored in this study are benign examples and do not involve algorithms used for generating adversarial machine learning samples. This study highlights the significance of enhancing the robustness of object detection models in real-life scenarios and the need for further investigation in this area to improve their accuracy and reliability.
\end{abstract}

\section{Introduction}
Traffic sign recognition is one significant use of these models, which has grown in importance as autonomous vehicles become more common. As autonomous vehicles become more widespread, the development of traffic sign recognition systems has received significant attention. The primary goal of traffic sign recognition systems in autonomous vehicles is to precisely recognize and categorize signs within digital images captured by sensors integrated into smart vehicles. The ability of Object Recognition (OR) models to recognize and classify these signals serves as the foundation for decision-making within the vehicle, directing various actions such as stopping at a stop sign. This demonstrates the critical role that traffic sign recognition plays in leading vehicle behavior based on interpreted images.

Among the various Object Recognition models, notable models are Faster R-CNN, SSD (Single Shot Detector), and YOLO (You Only Look Once). Within these, the YOLO model is particularly popular for traffic sign recognition applications. YOLO's strengths are its real-time processing capabilities which enables it to detect many objects fast and effectively in a single pass. Its ability to evaluate the complete image comprehensively leads to faster inference, resulting in it being suitable for applications that are dynamic like traffic sign recognition. Additionally, YOLO's ability to handle object dimension fluctuations while maintaining high detection accuracy even in crowded environment increases its effectiveness in such crucial field. However, despite the advancements of YOLO models in object detection their practical implementation faces challenges, especially when objects undergo alterations or manipulations.

Another popular study in this domain is to generate deceptive stickers that are designed to induce misclassification in these models. Our study presents an unconventional approach by focusing on benign altered examples where changes occur naturally. The purpose of this study is to investigate the YOLO model's robustness to real-world alterations, as opposed to techniques that intentionally manipulate inputs. This research is critical for understanding the model's resistance to changes caused by environmental conditions as well as providing critical insights to improve safety and reliability particularly in the area of autonomous vehicles that rely on these systems for making decisions.

In contrast to previous research, we refrain from creating adversarial examples created by adding perturbations generated utilizing mathematical algorithms to increase the YOLO model's misclassification rate.  Instead, our focus is on scrutinizing the model by testing it will benign cases that may reveal the sensitivity of the YOLOv7 model as YOLOv7 is regarded as the most efficient version among the YOLO variants. The study meticulously analyzes the model's performance in recognizing traffic signs under a range of scenarios, including variations in sign placement and adversarial deformations altering the shape, content, color, contrast and background of the traffic signs. Furthermore, the study evaluates the model's detection confidence under various angles and sign visibility conditions. A noteworthy finding is that the YOLOv7 model is more sensitive to traffic sign displacement which can be considered a serious vulnarability of the model. The evaluation section demonstrates the situation in which the model correctly detects traffic signs on human clothing with high confidence levels while raising concern on the model's overall resilience.

The contributions of this study can be summarized as follows-
\begin{itemize}
    \item Our study contributes insights into the practical challenges faced by YOLO models, especially in scenarios where objects, such as traffic signs undergo alterations or manipulations offering a nuanced understanding of the model's limitations.
    \item Our unique approach of introducing benign altered examples where modifications occur naturally provides a different perspective on assessing the YOLO model's robustness against real-world alterations which differs from intentional manipulations.
    \item Our research uncovers YOLO model’s sensitivity to traffic sign displacement and it becomes crucial for comprehending it’s resilience to alterations arising from environmental factors.
    \item Valuable insights offered by our study can lead a direction of study that can be helpful to enhance safety and reliability in the domain of autonomous vehicles relying on these systems for decision-making.
\end{itemize}

\section{Related Work}

Numerous research papers are exploring the creation of adversarial examples to fool the machine learning models. These papers investigate various methods for generating input images that can mislead the models into making mistakes. By crafting these malicious examples the researchers aim to highlight vulnerabilities in the reliability and adaptability of machine learning algorithms. Y. Li ~\etal in ~\cite{li2021adaptive} introduces a new attacking method called Adaptive Square Attack (ASA) which is designed for black-box attacks on recognition models. This attack employs an efficient sampling strategy to generate perturbations in traffic sign images demonstrating high efficiency in misclassifying state-of-the-art recognition models. Another work by B. Ye ~\etal proposes adversarial patch method which effectively attacks traffic sign recognition models and achieves high success rates on GTSRB-ResNet34 and GTSRB-GoogLeNet with an ensemble method enhancing the attack on GTSRB-VGG16. The crafted patches exhibit robustness across diverse transformations and random locations which shows over 70\% attack success rate in physical experiments ~\cite{9564956}. In ~\cite{pavlitska2023adversarial} S. Pavlitska ~\etal presents a survey conducting digital or real-world attacks on traffic sign detection and classification models, offering insights into recent advancements and identifying areas in research that need further exploration.

Several studies have also been done on traffic sign recognition using different versions of the YOLO model. J. Zhang~\etal displays real-time Chinese traffic sign detection utilizing the YOLOv2 model~\cite{intro1}. M. William ~\etal demonstrates the results of YOLOv2 on the German Traffic Signs Detection Benchmark (GTSDB) dataset~\cite{intro2}, whereas ~\cite{intro3} evaluates the performance of YOLOv3 on the same dataset. Aside from evaluating the model's performance, a lot of studies have been done by exploiting the vulnerabilities of the models and creating successful adversarial examples to degrade the model's performance. Among all the attack strategies, creating adversarial patches and adding adversarial perturbations to the images are the two most well-established research areas on object detection models. In recent times B. Ye~\etal proposes an adversarial patch method to attack the recognition of traffic signs on a similar dataset of GTSRB and the attack success rate (ASR) was reported over 90\% ~\cite{intro4}. N. Morgulis \etal discuss the use of adversarial traffic signs to attack neural-network-based classifiers like YOLO and present a pipeline for the reproducible production of such signs with the Fast Gradient Sign Method (FGSM) that can fool the classifiers, both open-source and production-grade in the real world. Moreover, ~\cite{intro5} explores the vulnerability of object detection models to physical adversarial examples, demonstrating a Disappearance Attack on Stop signs that fools YOLO v2 and Faster R-CNN in a controlled lab environment and outdoor experiment, and presents preliminary results of a Creation Attack using innocuous physical stickers.

Our approach differs from previous work as it does not focus on the creation of adversarial examples through mathematical perturbations to increase the YOLO model's misclassification rate. Instead, we focus on evaluating the YOLOv7 model's sensitivity by testing it with benign examples highlighting its robustness against examples that can occur spontaneously and can reveal the weakness of the model against those kind of samples. We analyze the model's behavior in recognizing traffic signs under multiple conditions like different placement of the traffic sign and adversarial deformation of a sign by altering its size, content, color, contrast and background. The study also evaluates the detection confidence of the model under various angles and the visibility of the sign. One of the noble findings of this research is that the YOLOv7 model is very much sensitive to the displacement of the traffic signs. The evaluation section demonstrates how a model detects traffic signs printed on the clothing of humans and these detections have a very high confidence level which clearly generates uncertainty about the model's robustness.

% \iffalse
% Finally from our experiments, we try to answer the following questions:
% \begin{enumerate}
%     \item What the minimal epsilon value could be for the maximal success rate of the FGSM attack in order to create adversarial perturbations that can make the model misclassify for each of the classes selected for testing.

%     \item What the optimal combination ratio could be for the regular images and FGSM-generated adversarial images that can be used as a training set to train a robust model that can defend against adversarial inputs with minimal performance degradation.

%     \item What the computational cost will be based on CPU/GPU power and time requirement in order to conduct the attack specified in RQ1 and in order to train the robust model specified in RQ2.

%     \item How robust the most current version of the YOLO OR model is to defend against the Daedalus attack, and what could be some ways to increase its robustness.
% \end{enumerate}
% \fi

\section{Background}

There are several popular object detection models, each with its own advantages and disadvantages. Based on the application, different models are selected for object detection purposes. The models can also be vulnerable to different adversarial attacks. This section discusses the proposed object detection model, its advantages, and also its vulnerability.

\subsection{YOLO}

YOLO is commonly used in computer vision applications. It was developed by Redmon~\etal in 2016. YOLO is a real-time object detection system that is capable of processing images extremely quickly. It works by dividing an image into a grid of cells, and each cell predicts the bounding boxes and class probabilities for the objects within it. The predictions from all the cells are combined to produce the final output of the model. Eqn~\eqref{yolo} shows the outputs of a YOLO model.

\begin{equation}
    Y = [p_c, b_x, b_y, b_h, b_w, c_1, c_2, . . , c_n]
    \label{yolo}
\end{equation}

Here, $ p_c $ stands for the probability score of the grid containing an object. $b_x, b_y $ are the x and y coordinates of the center of the bounding box with respect to the grid cell. $b_h, b_w $ corresponds to the height and the width of the bounding box with respect to the grid cell. $c_1, c_2, c_3, . . .c_n$ correspond to the class probabilities.

This shows that there will be multiple boxes of varying $p_c$ from which only one will be needed for identifying the object location. This is done in two steps: Bounding box regression and non-maximum suppression (NMS). The regression process keeps the significant boxes from all residual grids and the NMS gets rid of the boxes identifying the same object.

\subsection{YOLO in Traffic Sign Recognition}

Traffic sign recognition models are of critical importance in the development and operation of autonomous vehicles. Autonomous vehicles rely on a variety of sensors and models to navigate and make decisions, and traffic sign recognition models play a key role in this system. One of the main benefits of traffic sign recognition models in autonomous vehicles is improving safety. By detecting and interpreting traffic signs in real time, these models can help the vehicle make safe and efficient decisions on the road. By accurately detecting and interpreting traffic signs, these models can help the vehicle to optimize its route and speed, reducing travel time and energy consumption. Overall, traffic sign recognition models are critical for the safe and efficient operation of autonomous vehicles and are an important component of the sensor and model systems used in these vehicles.
Traffic sign recognition requires detecting and localizing signs in real time, which can be challenging due to the small size of the signs and the need for high accuracy. YOLO is designed to detect and classify objects in a single pass, which allows for real-time object detection and localization. This is especially useful for traffic sign recognition, as it can quickly detect signs and provide accurate location information. Additionally, YOLO uses anchor boxes to improve object detection, which helps to improve accuracy and reduce false positives. This is particularly important for traffic sign recognition, as false positives can lead to incorrect or dangerous decisions being made by autonomous vehicles or drivers. YOLO has also been shown to achieve high accuracy and speed on a variety of object detection tasks, including traffic sign recognition~\cite{traffic_yolo}. This makes it a popular choice for traffic sign recognition systems used in autonomous vehicles, where accuracy and speed are critical for safe and efficient operation.

\subsection{Traffic Sign Recognition Alteration}

Object recognition models rely on a set of features or patterns in an image to identify and classify objects accurately. These features can be geometric, statistical, or textural, and they capture unique characteristics of objects in the image. When these features are altered, either intentionally or unintentionally, the object recognition model may fail to detect and classify objects correctly. For instance, changes in size, shape, and angles can alter the geometric features of objects, such as their aspect ratio, symmetry, or position, making it difficult for the model to recognize them. Similarly, changes in color or visibility can alter the statistical and textural features of objects, such as their brightness, contrast, or texture, making them less distinguishable from the background or other objects in the image. Mathematically, these changes can affect the input features of the object recognition model, which in turn, can affect the model's output. For example, changes in size and shape can alter the size and position of the bounding boxes used to detect objects, leading to incorrect predictions. Changes in color or visibility can affect the pixel values of the image, leading to incorrect feature extraction and classification. Therefore, it is essential to develop object recognition models that are robust to these types of alterations to ensure their reliability and accuracy in real-world scenarios.

The feature extractor is usually implemented using a deep convolutional neural network (CNN), which is trained on a large dataset of labeled images. The CNN learns a set of filters or kernels that convolve with the input image to extract features at different scales and orientations as Eqn~\eqref{CNN} where x is the input image, w is the filter, b is the bias, and the * symbol represents convolution. The output of the convolution operation is a feature map that captures the presence of the filter in different regions of the image.

 \begin{equation}
    x * w = \sum_{i=1}^{n} x_iw_i + b
    \label{CNN}
\end{equation}

The classifier is typically implemented using a fully connected layer, which takes the features extracted by the CNN and outputs a probability distribution over the classes. Mathematically, this can be represented as Eqn~\eqref{CNN2} where x is the output of the CNN, W is the weight matrix, b is the bias, and pi is an activation function, such as ReLU or SoftMax. The output y is a vector of probabilities over the classes.

 \begin{equation}
    y = \phi (Wx+b)
    \label{CNN2}
\end{equation}

Changes in size, shape, color, visibility, and angles can affect the output of the feature extractor and, consequently, the performance of the classifier. For example, changes in size and shape can affect the receptive field of the filters, leading to a loss of spatial resolution or a failure to detect objects at certain scales or positions. In Eqn~\eqref{CNN3} yij is the output of the jth filter at position (i,j), xkl is the input at position (k,l), wkl is the weight of the filter, and bj is the bias.

 \begin{equation}
    y = \phi (\sum_{kl} X_{k+i-1 , l+j-1}W_{kl}+b_j )
    \label{CNN3}
\end{equation}

Changes in color or visibility can affect the distribution of pixel values in the image, leading to incorrect feature extraction and classification. Eqn~\eqref{CNN4} can represent the distribution where x is the input image, m is a mask that represents the changes in color or visibility. The output y is a vector of probabilities over the classes.

 \begin{equation}
    y = \phi (W(x \otimes m) +b )
    \label{CNN4}
\end{equation}

\vskip 0.15in
\section{Approach}
\label{sec:apprch}
This section discusses the approaches for evaluating each condition to reveal the sensitivity of the YOLOv7 model:

\subsection{Displacement}
Different traffic signs were printed on human clothing to assess the model's performance. Moreover, printed traffic signs were placed in various irrelevant places where an original traffic sign should not be placed and detected like behind another vehicle or in the middle of the road. These scenarios are considered to reveal that a robust model should be able to detect the traffic signs only in the places it should be detected.

\subsection{Non-adversarial stickers}
As various symbols are used by vehicles like trucks to warn nearby drivers and vehicles, it is interesting to see if such symbols or stickers can impact the model's recognizing ability by misguiding the model to detect unrelated stickers or symbols as traffic signs.

\subsection{Deformation}
The models' sign recognition capability was also evaluated by changing each traffic sign's shape, size, and color. The prepared adversarial traffic signs contain all the possible colors from the RGB panel to justify the model's robustness. In the case of shape, each sign was altered with the shape of another like a stop sign being round like a speed-limit sign with a regular color combination. Moreover, various sizes of traffic signs were tested ranging from 0x to 100x zoomed to demonstrate how the model reacts to way too large signs and way too small signs.

\subsection{Natural Conditions}
Testing the recognition ability and confidence of the model on traffic signs in various environmental conditions such as rain, snow, and mud to see the robustness of the models is important as the model's ability to recognize traffic signs in such conditions might be the only thing between an autonomous vehicle being involved in an accident or not. We plan to test the ability of the YOLO model by checking how well it recognizes traffic signs in various natural deformations like being covered in snow, high exposure, and lighting condition.

\vskip 0.15in

\section{Experimental Setup and Training}

\begin{figure}[htbp]
\centering
\includegraphics[width=\columnwidth]{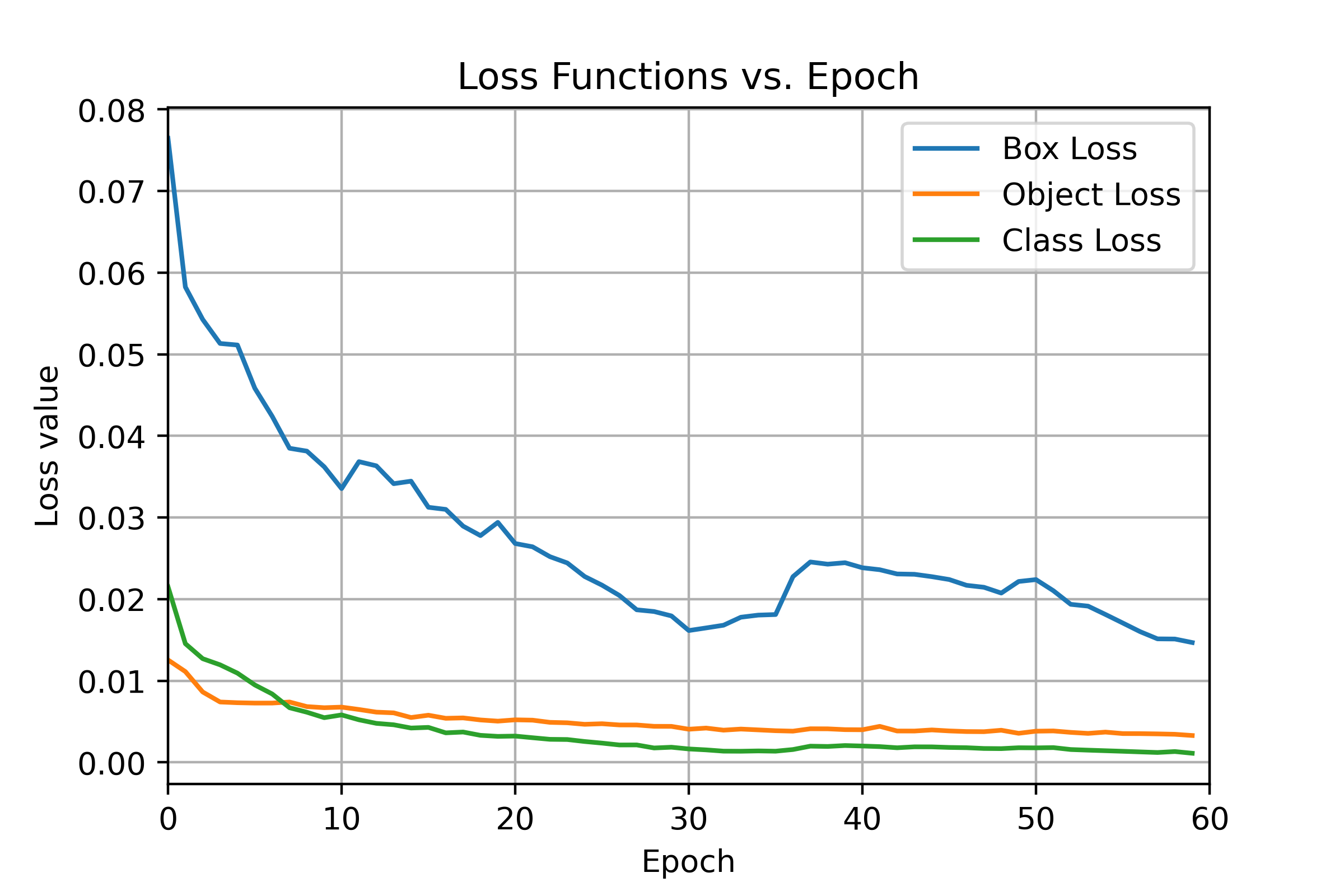}
\caption{Epoch vs loss}
\end{figure}

\begin{figure}[htbp]
    \centering
\includegraphics[width=\columnwidth]{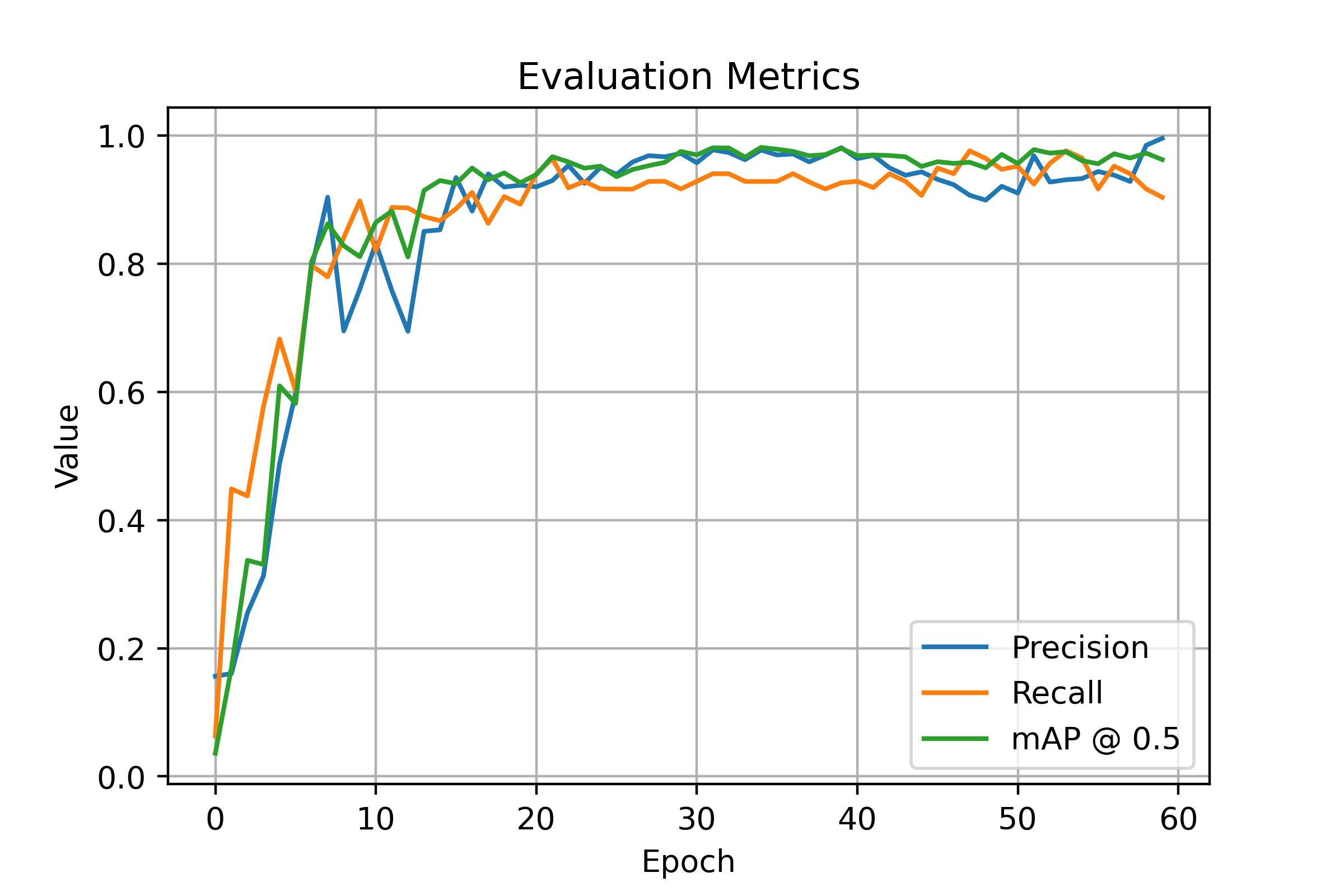}
  \caption{Epoch vs Metrics}
\end{figure}

Upon investigating such two-pass architectures as HOG and Faster R-CNN and single-pass architectures such as SSD and YOLO, our team chose to investigate the YOLO architecture as it appeared to be the most supported and state-of-the-art detecting model. Newer versions of YOLO boast capabilities of processing and labeling photos at upwards of 500 frames per second and have many industries' use cases, increasing the likelihood that this investigation may have real-world impacts.

In determining a YOLO version to work with, we chose between YOLOv5, YOLOv7, and YOLOv8. While there were more resources available for YOLOv5, sufficient information existed to train and implement YOLOv7~\cite{yolov7}. We chose not to use YOLOv8 due to its relatively recent roll-out, approximately 3 months at the time of writing, due to potential instabilities or bugs. Careful consideration must be made with respect to the dataset used for training in order to keep the application realistic and impactful. Perturbing houseplant identification, while still useful, would not be as serious an issue as attacking everyday logistic items. We began by training a YOLOv5 model on a German road sign dataset~\cite{make_ml} – This was the most pragmatic choice as it was an easily found tutorial on how to train the models as well as align with project goals.

Further steps were taken towards independence by investigating a Facemask dataset~\cite{make_ml1}, which required unguided implementation of the YOLOv5 architecture. A final move to YOLOv7 undertook an investigation of a third dataset that allowed training on European traffic signs~\cite{make_ml2} with the following four (4) classes:

\begin{itemize}
    \item Traffic lights
    \item Stop signs
    \item Pedestrian crossings
    \item Speed limit signs
\end{itemize}

This step involved advancements in understanding the AI architecture such as label conversion (PASCAL VOC to YOLO format) and partitioning of training, validation, and test sets. Datasets such as this with common traffic signs allow us to quantify the very real dangers of the weaknesses posed by deep neural networks whose hands we may put our lives in during autonomous driving, for instance.

Initial performance looks good on this dataset for the test data, showing high values for precision, recall, and mean average precision (mAP) at 50\% IoU. These results can be seen in Table~\ref{result}. It should be noted that each of these training began with supplied pre-trained weights found on the architectures’ respective GitHub, not with random or unweighted starting points.

\begin{table}[htbp]
\caption{Summary of European traffic signs training dataset using YOLOv7 architecture}
    \centering
    \resizebox{\columnwidth}{!}{
    \setlength{\tabcolsep}{10pt}
\begin{tabular}{cccccc}
\toprule
\bf Class & \bf Images & \bf Labels & \bf P & \bf R & \bf mAP@.5 \\
\midrule
all & 88 & 136 & 0.972 & 0.917 & 0.975 \\
 traffic light & 88 & 21 & 0.941 & 0.762 & 0.926 \\
speed limit & 88 & 85 & 0.997 & 1 & 0.996 \\
 crosswalk & 88 & 21 & 0.953 & 0.905 & 0.984 \\
stop & 88 & 9 & 0.997 & 1 & 0.995 \\
\bottomrule
\end{tabular}
}
\label{result}
\end{table}

The datasets evaluated thus far are certainly impactful. As shown in Figure~\ref{fig:graph} the most recent training results for our European road signs were as desired and these results are for training on 60 epochs.

\section{Results}
To understand how the decision-making process works in the YOLO model, we tried to see how the model classifies objects and recognizes them based on the following features:

\begin{itemize}
  \item How changing shape affects the model's recognition
  \item How changing color affects the model's recognition
  \item How adding natural patches or textures like snow affects the model's recognition
  \item How the distance and angle play a role in the recognition ability of the models
\end{itemize}

\subsection{Displacement}
The performance of the models was assessed by printing various traffic signs on human clothing, and by placing printed traffic signs in various irrelevant locations where an original traffic sign should not be located, such as behind another vehicle or in the middle of the road. These scenarios were designed to demonstrate that a robust model should only detect traffic signs in their designated locations.

\subsection{Shape}
\begin{figure*}
\centering 
\subfloat[Green-colored sign detected as Stop sign]{
    \includegraphics[width=.28\textwidth]{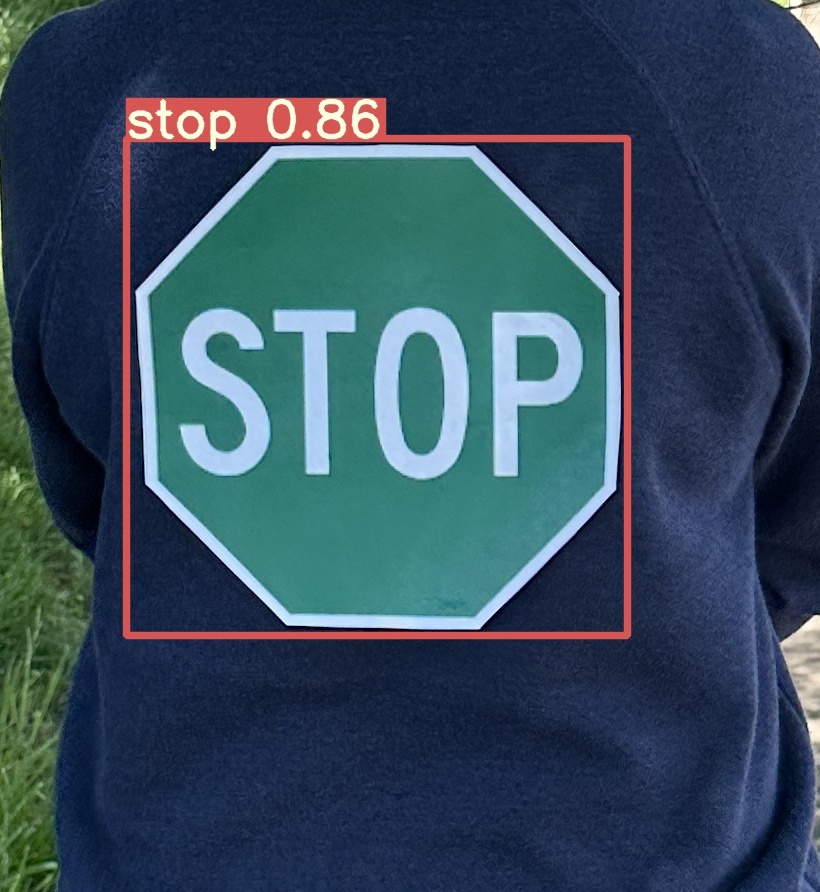}
    \label{fig:green_stop_sign}
    }
    \subfloat[Stop sign shape with KFC text]{    \includegraphics[width=.3\textwidth]{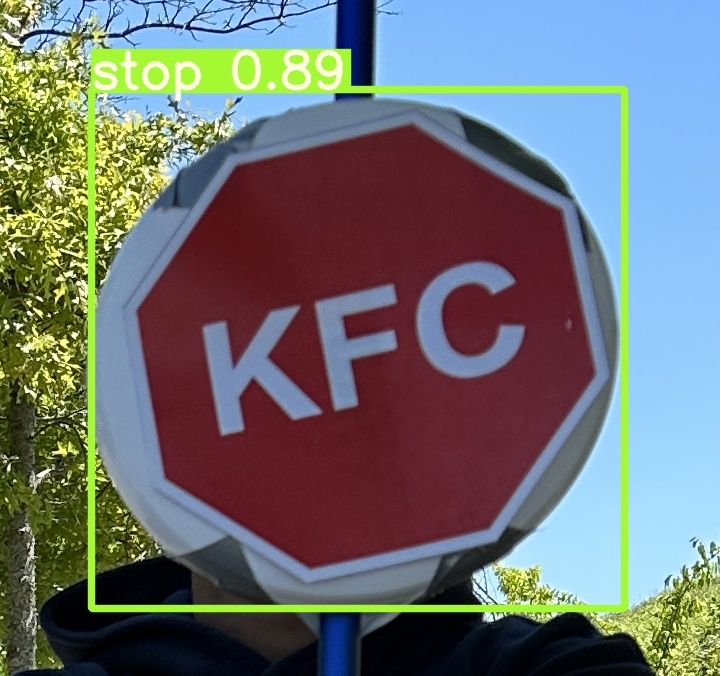}
    \label{fig:KFC1}
    }
\subfloat[Stop sign shape with KFC text]{
    \includegraphics[width=.3\textwidth]{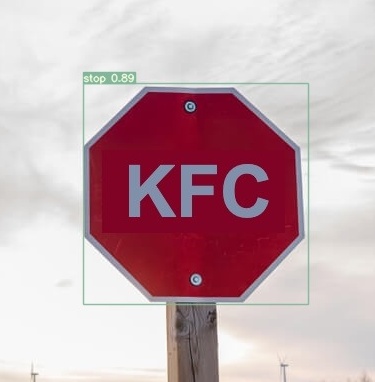}
    \label{fig:KFC2}
    }
\end{figure*}

Regarding the four categories of traffic signs selected in our research work, some have no effect on changing shape. However, traffic signs like the Stop sign and the Speed sign seem to be affected by changing shape. In ~\ref{fig:KFC1}, and ~\ref{fig:KFC2}, both of them had the same shape and color as a stop sign. The model recognized them as stop signs with more than 80\% confidence. Moreover, in ~\ref{fig:green_stop_sign}, the octagonal-shaped green STOP symbol has also been recognized as a Stop sign with more than 80\% confidence by the YOLO model emphasizing the importance of shape as a core feature that the model uses for classification.

\begin{figure*}
\centering 
    \includegraphics[width=1.5\columnwidth]{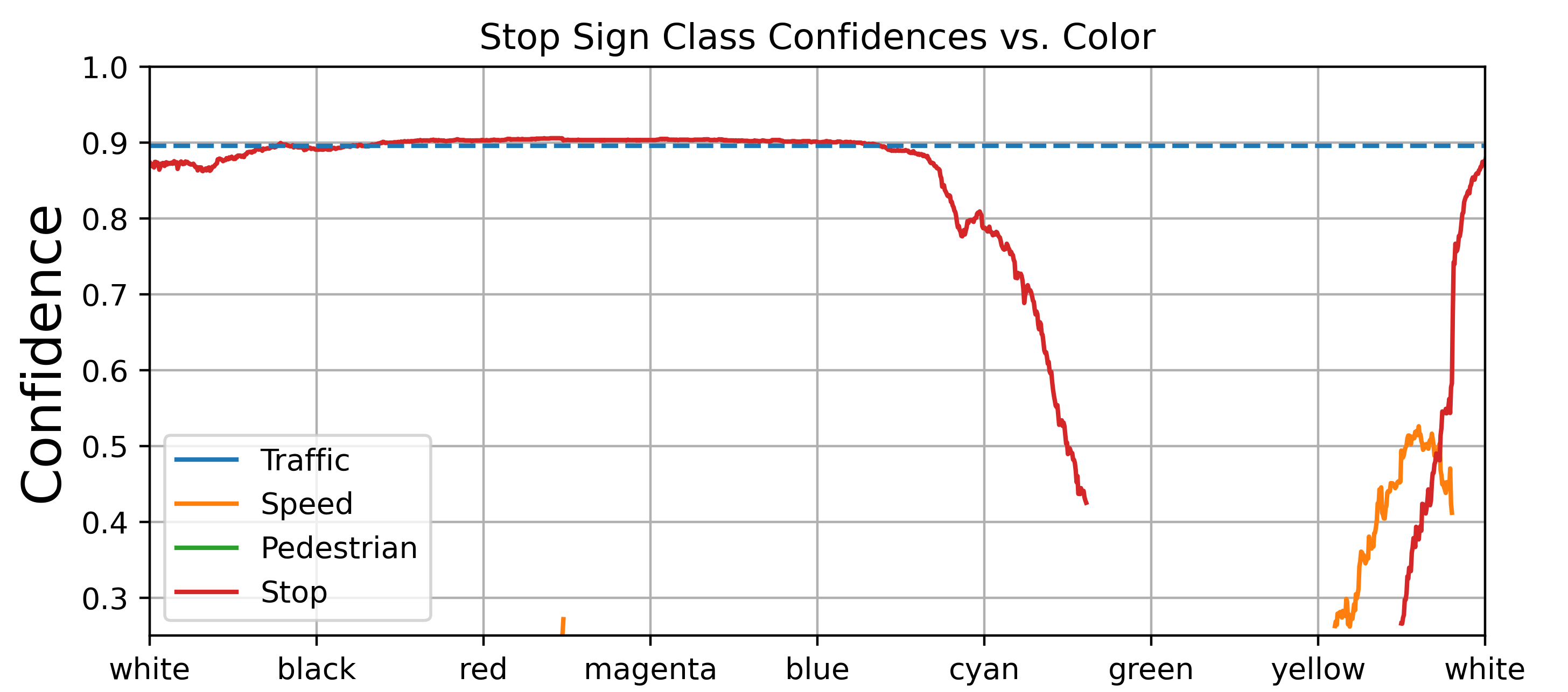}
    \caption{Graph representation of model's confidence by changing color channels}
    \label{fig:ColorGraph}
\end{figure*}

\begin{figure}
    \includegraphics[width=0.8\columnwidth]{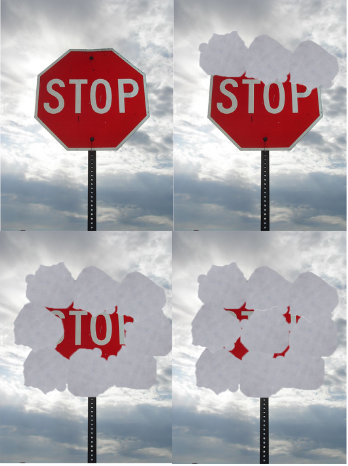}
    \caption{Snow-covered stop sign}
    \label{fig:stop_in_snow_pic}
\end{figure}

\begin{figure}
    \includegraphics[width = \columnwidth]{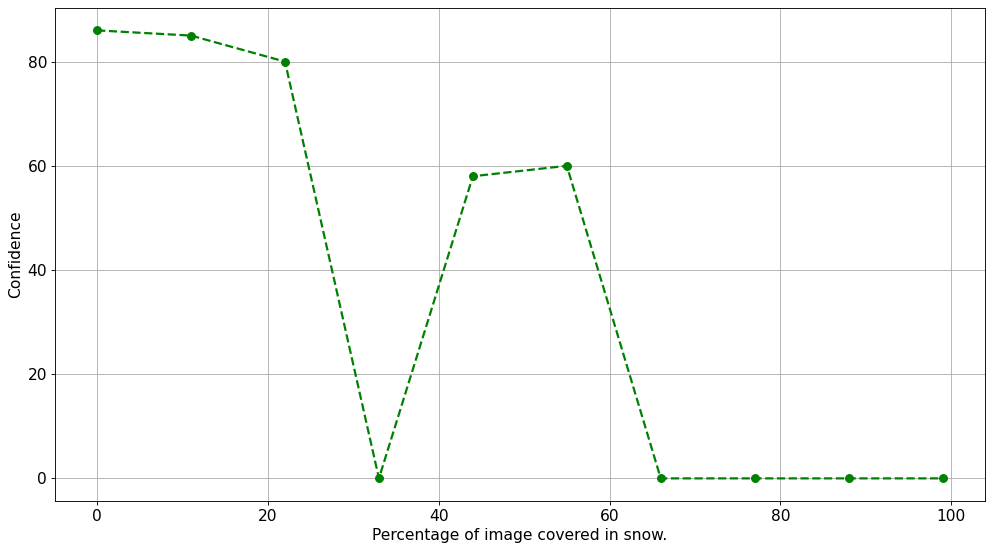}
    \caption{Graph reflecting the confidence change when various parts of the stop sign are covered in snow patches.}
    \label{fig:stop_in_snow_graph}
\end{figure}

\begin{figure}
\centering
    \includegraphics[width = \columnwidth]{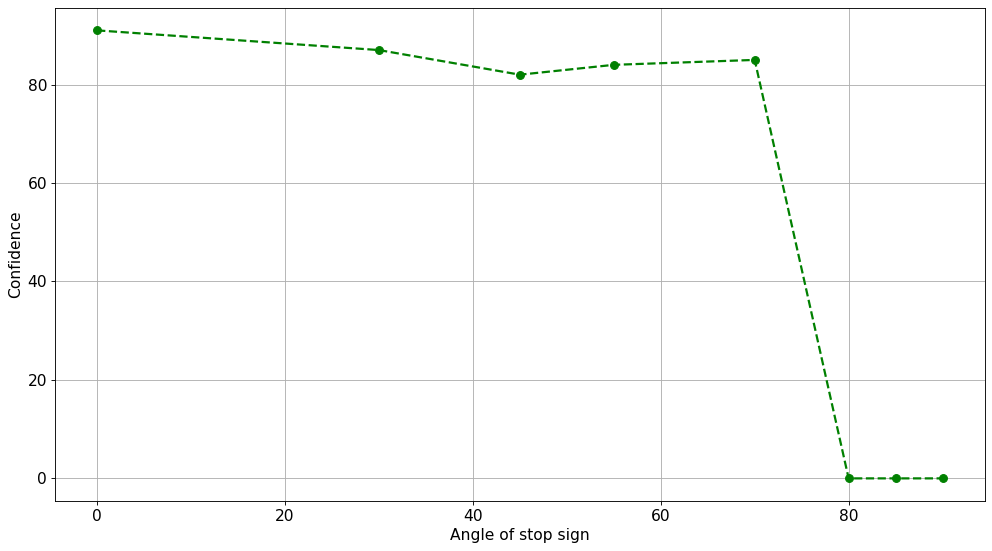}
    \caption{Graph reflecting the confidence change when the stop sign front is at various angles relative to the viewing.}
    \label{fig:stop_in_angle}
\end{figure}

\subsection{Color}
Looking at the results from our previous experiments, the YOLO seems to classify octagonal-shaped red objects as stop signs as those two are the primary feature of a stop sign. As the training set reflects the traffic light having various colors, the model always shows high confidence in detecting traffic lights of various colors. On the other hand, the speed sign, and stop sign confidence changes based on various background colors. The highest drop is reflected in these two signs when the red channel value is zero or moves toward zero. This is reflected in ~\ref{fig:ColorGraph}, graph.

\begin{figure}[htbp]
\centering
    \includegraphics[width=\columnwidth]{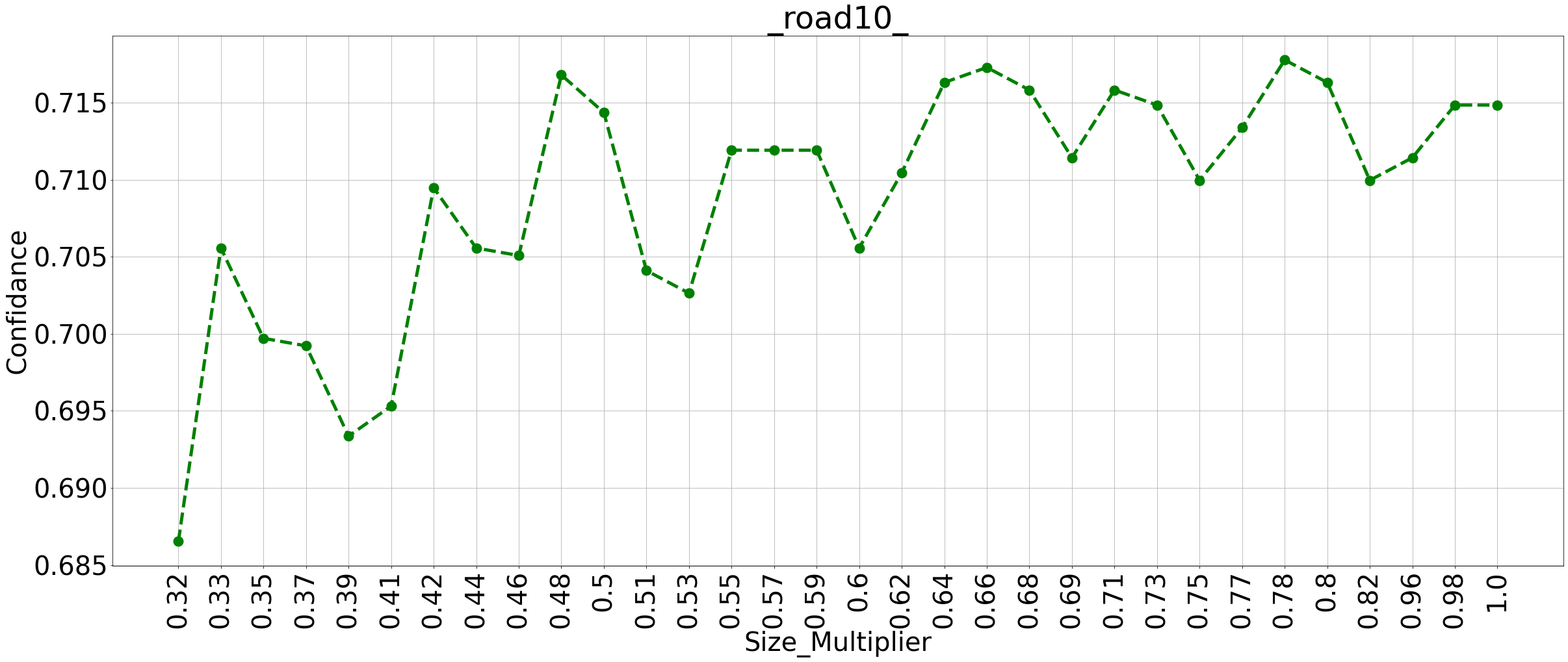}
    \caption{Size Multiplier vs Confidence for road10}
    \label{fig:_road10_}
\end{figure}

\begin{figure}[htbp]
    \centering   
    \includegraphics[width=\columnwidth]{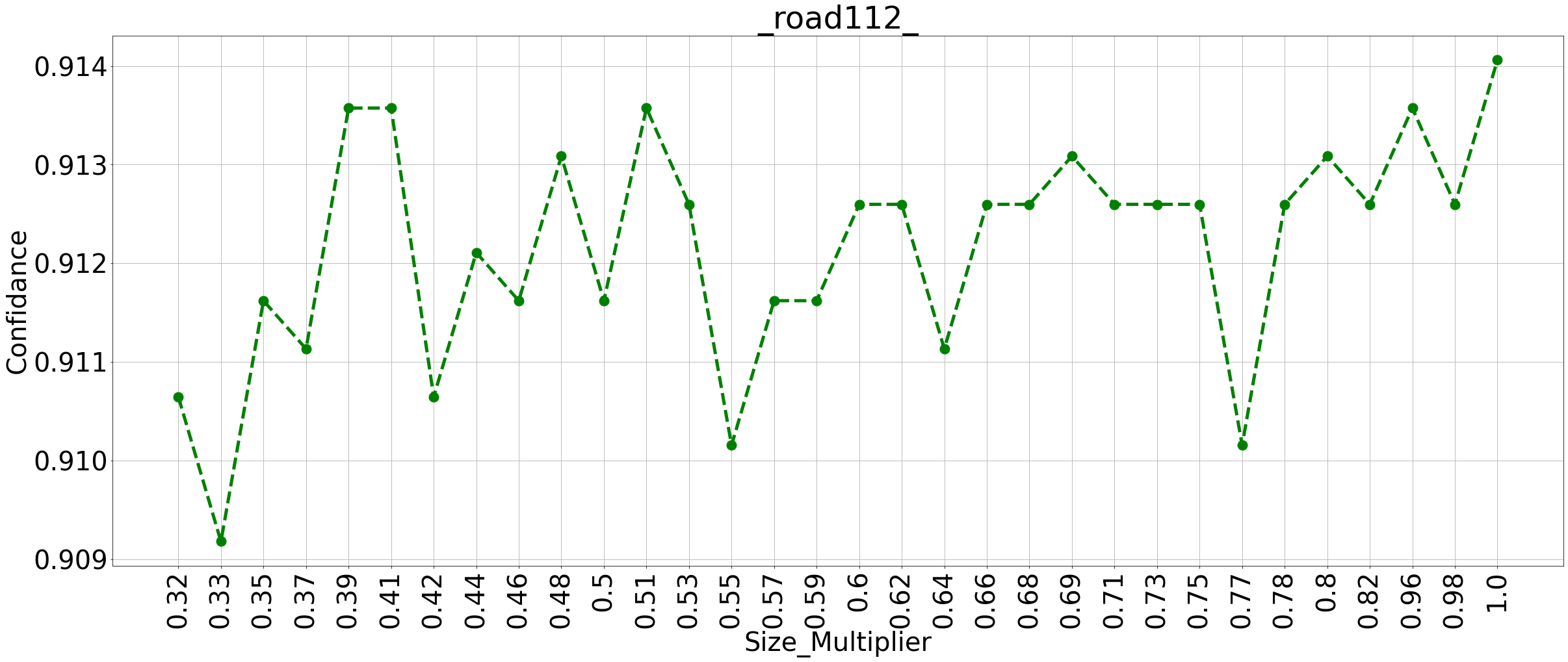}
    \caption{Size Multiplier vs Confidence for road112}
    \label{fig:_road112_}
\end{figure}

\subsection{Natural Patches}
To reflect natural patches, we added known patterns on stop signs to see how well the YOLO model detects the stop sign based on various percentages of the stop sign being covered by snow-like patches. The graph in ~\ref{fig:stop_in_snow_graph} shows how the stop sign detection confidence drops as more parts of the stop sign are covered by snow. However, one special condition can be seen when the model cannot detect the stop sign when the upper octagonal part of the stop sign shape is not properly seen even with relatively less part of the entire stop sign being covered by snow patches. However, adding the same amount of snow on the bottom part of the image doesn't result in the same outcomes.

\subsection{Angle relative to the viewing camera}
Even when the stop sign's front is faced at a greater than 45-degree angle, the YOLO model can recognize the stop sign as a stop sign with high confidence. Only after crossing a 70-degree angle, the model starts to stop recognizing the stop sign. The graph in Figure 7 shows the model's confidence relative to various angles of the stop sign front.

\section{Conclusion}
Traffic sign recognition is a crucial task in various applications such as autonomous driving, as it enables vehicles to navigate safely and efficiently on the roads. The findings of this analysis provide important insights into the factors that influence the performance of popular traffic sign recognition models such as the YOLOv7 model, which can ultimately impact the safety of passengers and other road users. 

The study reveals that the color, shape, and content of traffic signs are all important features that affect the model's confidence in traffic sign recognition. The study also demonstrates that changes in color, shape, and viewing conditions can drastically impact the recognition ability of the model. Although the YOLOv7 model is robust in detecting traffic signs events in various hard environmental conditions, such as snow-covered traffic signs and high exposure. Our study also highlights the challenges of quantifying the effects of changes in the model's  detection performance. While changes in content can have an impact on the model's confidence, the study suggests that these effects are difficult to quantify. This highlights the need for further research in this area to better understand the factors that influence traffic sign recognition performance.

In summary, this analysis provides valuable insights into the factors that influence the performance of traffic sign recognition models. By highlighting the importance of color, shape, and content, as well as the trade-offs associated with these features, this study provides important guidance for the design and deployment of traffic signs in the real world to safely operate autonomous vehicles. These findings underscore the importance of robust training and testing to ensure reliable performance in real-world scenarios, ultimately contributing to safer and more efficient autonomous driving.

\section{Future Work}
One of the greatest struggles we faced in this research is the lack of a good US traffic sign dataset. Therefore, all our training has been conducted on a European traffic sign dataset. Therefore, the discrepancy in traffic signs between Europe and the US has forced us to use a limited number of traffic signs to test. In the future, we plan to train a model using US traffic signs to see if the performance of the model has any changes due to using different shapes or colors for different categories of signs. Furthermore, our research was limited to four distinct traffic signs. More work needs to be conducted to see if traffic signs like various speed limits and similar shaped signs confuse the model or not. A few interesting questions came to light while summarizing the findings of this research. The first, question is, does the model prioritizes certain parts of the traffic signs more in making decisions, if so, which parts are they, and can training the model by modifying those parts make the model more robust? Furthermore, how well does the model handle different-looking same types of traffic signs from various parts of the world? More work needs to be done to answer these questions.

\bibliographystyle{IEEEtran}
\bibliography{reference}\balance

\end{document}